\documentclass{article}

\usepackage{microtype}
\usepackage{graphicx}
\usepackage{subfigure}
\usepackage{booktabs} 

\usepackage{hyperref}

\usepackage[accepted]{icml2024}

\usepackage{amsmath}
\usepackage{amssymb}
\usepackage{mathtools}
\usepackage{amsthm}

\usepackage[capitalize,noabbrev]{cleveref}

\theoremstyle{plain}

\theoremstyle{definition}

\theoremstyle{remark}

\usepackage{multirow}
\usepackage{color}
\usepackage{bm}
\usepackage{booktabs}
\usepackage{bbding}
\usepackage{hyperref}
\usepackage{algorithm}  
\usepackage{array}   
\usepackage{colortbl} 
\usepackage{cleveref}  

\definecolor{myred}{RGB}{210, 10, 10}
\definecolor{mygreen}{RGB}{10, 210, 10}

\definecolor{lightgray}{rgb}{0.9, 0.9, 0.9}

\definecolor{baselinecolor}{gray}{.9}

\usepackage{pifont} 
\usepackage[textsize=tiny]{todonotes}

\icmltitlerunning{Feast Your Eyes:  Mixture-of-Resolution Adaptation for Multimodal   Large Language Models}

\begin{document}

\twocolumn[
\icmltitle{Feast Your Eyes:  Mixture-of-Resolution Adaptation for Multimodal \\ Large Language Models}



\icmlsetsymbol{equal}{*}

\begin{icmlauthorlist}
\icmlauthor{Gen Luo}{mac,pcl}
\icmlauthor{Yiyi Zhou}{mac,xmu}
\icmlauthor{Yuxin Zhang}{mac,xmu}
\icmlauthor{Xiawu Zheng}{mac,xmu}
\icmlauthor{Xiaoshuai Sun}{mac,xmu}
\icmlauthor{Rongrong Ji}{mac,xmu} 
\end{icmlauthorlist}

\icmlaffiliation{mac}{Key Laboratory of Multimedia Trusted Perception and Efficient Computing, Ministry of Education of China,  School of Informatics, Xiamen University, 361005, P.R. China}
\icmlaffiliation{xmu}{Institute of Artificial Intelligence, Xiamen University, 361005, P.R. China}
\icmlaffiliation{pcl}{Peng Cheng Laboratory, Shenzhen, 518000, China}

\icmlcorrespondingauthor{Rongrong Ji}{rrji@xmu.edu.cn} 


\vskip 0.3in
]



\printAffiliationsAndNotice{}  

\begin{abstract}
Despite  remarkable progress, existing multimodal large language models (MLLMs) are still inferior in granular visual recognition. Contrary to previous works, we study this problem from the perspective of image resolution, and reveal that a combination of low- and high-resolution visual features can effectively mitigate this shortcoming.  Based on this observation, we propose a novel and efficient method for MLLMs, termed \emph{Mixture-of-Resolution Adaptation} (MRA). In particular, MRA adopts two visual pathways for  images with different resolutions, where  high-resolution visual information is embedded into the low-resolution pathway via the novel \emph{mixture-of-resolution adapters} (MR-Adapters). This design also   greatly  reduces the input sequence length of MLLMs. To validate MRA, we apply it to a recent MLLM called LLaVA, and term the new model  \textit{LLaVA-HR}. We conduct extensive  experiments on 11 vision-language (VL) tasks, which show that LLaVA-HR outperforms existing MLLMs on 8 VL tasks, \emph{e.g.,} +9.4\% on TextVQA.  More importantly,    both training and inference  of LLaVA-HR remain efficient with MRA, \emph{e.g.,}  \textbf{\textit{20 training hours}} and  \textbf{\textit{3$\times$ inference speed}}  than LLaVA-1.5. Source codes are released at: \url{https://github.com/luogen1996/LLaVA-HR}.
 
\end{abstract}

\section{Introduction}
\label{submission}
 Driven by the remarkable success of large language models (LLMs)~\cite{llama,gpt3}, research on multi-modal large language models (MLLMs) also receives an influx of interest in the machine learning community~\cite{llava,lavin,alayrac2022flamingo,chen2022pali,chen2023pali}.   Numerous efforts have been recently devoted to extending LLMs to more modalities, achieving  breakthroughs on various vision-language tasks~\cite{goyal2017vqav2,singh2019textvqa,hudson2019gqa}. Despite    advances,  existing MLLMs still fall short of granular  visual  recognition. For instance,  the  powerful GPT4-V also suffers from hallucinations when identifying  small and occluded objects~\cite{visshortcoming}. This shortcoming inevitably limits the practical use of MLLMs.

	\begin{figure}[t]
		\centering
		\includegraphics[width=0.45\textwidth]{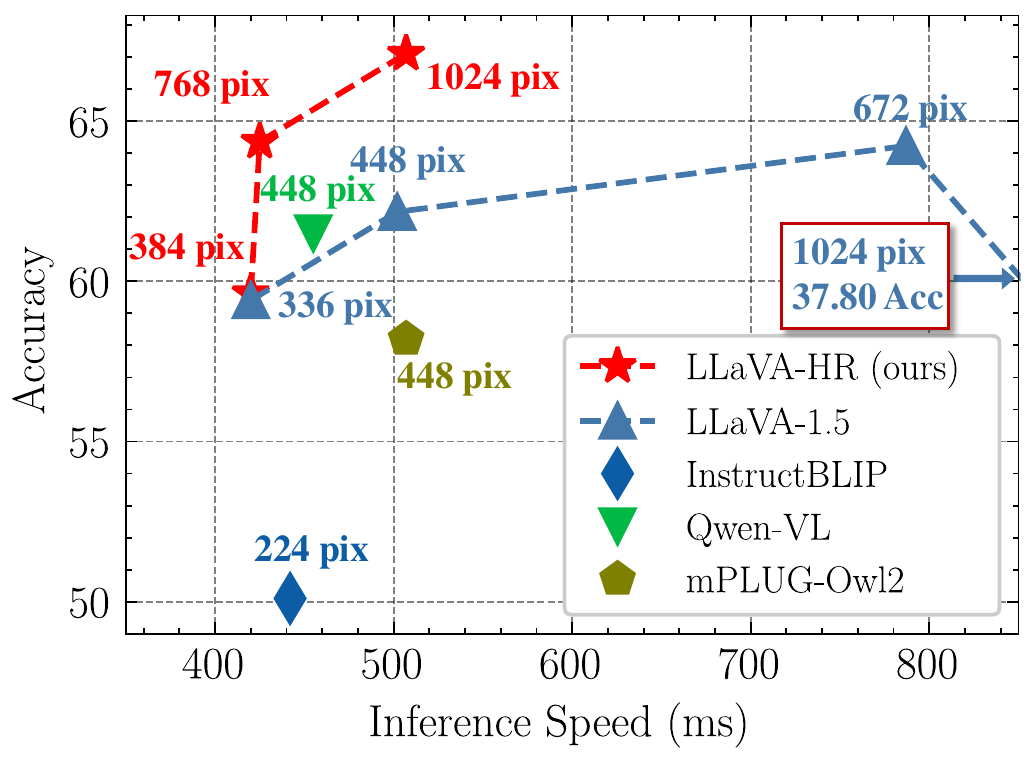} 
  		\vspace{-1em}
		\caption{\textbf{Zero-shot performance and inference speed of LLaVA-HR and existing MLLMs on TextVQA.}  Existing MLLMs often fall short  of fine-grained VL tasks like TextVQA.  Increasing   image resolution  is  an effective yet expensive solution. With the proposed MRA, our LLaVA-HR can efficiently adopt high-resolution images to boost performance.}
		\label{fig1}
		\vspace{-1.5em}
	\end{figure}

 \begin{figure*}[t]
		\centering
		\includegraphics[width=1.\textwidth]{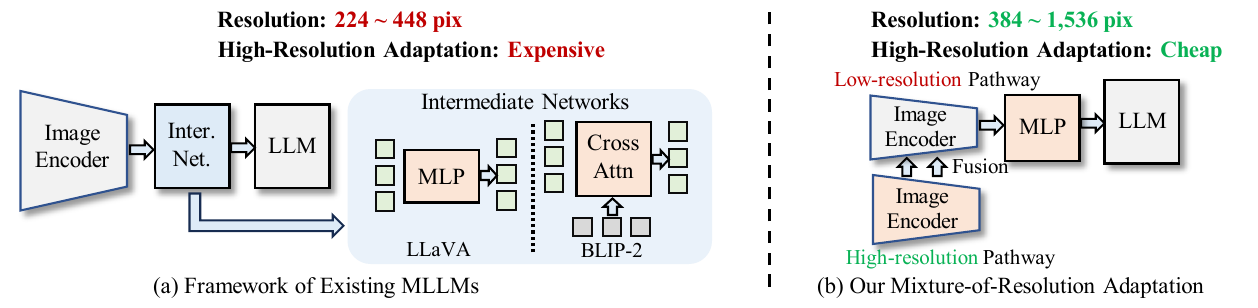} 
  		\vspace{-2.5em}
		\caption{\textbf{Comparison  between existing MLLMs and LLaVA-HR.}  Due to high computation complexity, existing MLLMs~\cite{llava1.5,li2023blip} often use input images of low-resolution, which are insufficient for granular visual reasoning.  With  our mixture-of-resolution adaptation,  the proposed  LLaVA-HR can increase the image resolution up to 1,536 $\times$ 1,536 with  limited additional costs. }
		\label{fig1-1}
  		\vspace{-1.em}
	\end{figure*}

  To compensate for this shortcoming, practitioners often resort to scaling up   model size  and increasing per-training data  size~\cite{alayrac2022flamingo,li2023blip,bai2023qwen}.   For instance, InstructBLIP~\cite{dai2023instructblip} adopts over 129M image-text pairs for vision-language (VL) alignments,  and shows   that  a  larger visual encoder  is beneficial for  MLLMs. Motivated by this, Qwen-VL~\cite{bai2023qwen} further increases the parameters of visual encoder to 1.9 billion and  uses 1.5 billion pre-training data.   Despite   progress,   this paradigm is prohibitively expensive,  which  often 
 consumes  about   thousands of  GPU hours.

Orthogonal to these works, we study the visual shortcoming of MLLMs from the perspective of input image resolutions.  As revealed in previous VL research~\cite{indefense,visshortcoming,simrec}, increasing the resolution of input images is a straightforward solution to improve  visual recognition, which becomes more important for  MLLMs that involve \textit{visual chain-of-thought}~\cite{rose2023visual}.   As shown in Fig.~\ref{fig1},  increasing the resolution  of LLaVA-1.5~\cite{llava1.5}  from 384 $\times$ 384 to 672 $\times$ 672 can bring obvious performance gains (+4.6\%)  on TextVQA~\cite{singh2019textvqa}.   However, the use of high-resolution images will greatly exacerbate the already high computational cost of MLLMs. For instance, $448\times448$ resolution will increase the computation complexity of LLaVA by about 1.4 times compared with the default $336\times 336$.  In addition,  due to the complex structure of MLLMs, the training will become unstable as the resolution is greatly increased, \emph{e.g.}, a sharp drop at $1,022\times 1,022$ resolution,  as shown in Fig.~\ref{fig1}. We assume that the length of visual sequences greatly exceeds the pre-trained context length, leading to training instability.

In this paper, we propose  a novel and efficient method  for the high-resolution image adaptation of MLLMs, namely \textit{mixture-of-resolution adaptation} (MRA). As shown in Fig.~\ref{fig1}, MRA adopts an innovative dual visual pathway design to process the input images of high- and low-resolutions simultaneously.  
Specifically,  one pathway aims to encode global information of low-resolution images, while the other  one serves to  capture  fine-grained semantics from high-resolution images.   Meanwhile, these two pathways are   closely interacted  via the novel \textit{mixture-of-resolution adapters} (MR-Adapters),  which embeds the high-resolution visual information into the low-resolution modeling. In this way, we can use a much fewer number of visual tokens to represent the input images from macro- to micro-views.  With the careful design of dual-pathway structure, MRA can  easily increase the image resolution  up to 1,536 $\times$ 1,536 pixels while maintaining high  efficiency.

To validate MRA, we apply it to a recent MLLLM called LLaVA~\cite{llava,llava1.5}, and term the new model as   LLaVA-HR.   We conduct extensive experiments on 11 vision-language (VL) tasks, including common VL tasks like VQA2.0~\cite{goyal2017vqav2}  and emerging   benchmarks such as POPE~\cite{li2023pope}.  Experimental results show that LLaVA-HR  outperforms existing MLLMs   on 8 of 11 VL tasks, \emph{e.g.,} +9.6\% over LLaVA-1.5 on  TextVQA.  More importantly,  the training and inference   of LLaVA-HR  are  cost-effective.   The pre-training and instruction tuning of LLaVA-HR (7B, 1,024 $\times$ 1,024) only take a total of 20.7 hours on 8 A800 GPUs, which is \textbf{\textit{hundreds of times cheaper}} than InstructBLIP~\cite{dai2023instructblip} and Qwen-VL~\cite{bai2023qwen}.  With the same  resolution, its inference speed  is  \textbf{\textit{3 times faster}} than LLaVA-1.5~\cite{llava1.5}.

 In summary, our contributions are three folds:
\begin{itemize}
    \item We reveal the significance of image resolution for MLLMs and propose a novel and efficient adaptation scheme, termed \emph{mixture-of-resolution adaption} (MRA), which adopts a novel dual visual pathway design to obtain the benefits of high-resolution visual information while keeping   training and inference efficient.
    \item We propose a novel\textit{ mixture-of-resolution adapter} (MR-Adapter) for MRA, which can  embed the high-resolution information into the low-resolution visual pathway  to improve  visual  descriptive power.
    \item Based on MRA, we propose a powerful MLLM,  coined  LLaVA-HR, which   outperforms existing MLLMs on 8 of 11 VL tasks and spends much cheaper training expenditure than most MLLMs.
\end{itemize}

\section{Related Work}
\subsection{Multimodal Large Language Models}
 Driven by the great successes of large language models (LLMs)~\cite{gilardi2023chatgpt,llama,gpt3},  growing interest has been aroused in building end-to-end multimodal large language models (MLLMs)~\cite{llava,zhu2023minigpt,lavin,fuyu,peng2023kosmos,liu2023llavaplus}. In particular, most existing MLLMs adopt a modular structure~\cite{lavin,llava}, which utilizes an intermediate network to project the visual features into the word embedding space of the LLM. Then, the LLM is used to accomplish various VL tasks in an autoregressive manner.   Based on the modular structure,  existing MLLMs can be distinguished by the designs of the intermediate network.   Popular MLLMs  represented by LLaVA~\cite{llava} often adopt a  linear projection layer or an MLP layer to connect the visual encoder and the LLM~\cite{llava,llava1.5,chen2023shikra,chen2023pali,peng2023kosmos}. The other works employ   sampler-based modules to bridge  the gap between the visual encoder and the LLM~\cite{bai2023qwen,alayrac2022flamingo,li2023blip,dai2023instructblip}. These sampler-based modules can effectively reduce the number of visual tokens, but often requires a large-scale pre-training to achieve a promising performance~\cite{bai2023qwen,li2023blip}.    Despite the effectiveness, most existing MLLMs still  adopt a low visual resolution, \emph{e.g.,} 336 $\times$ 336, which greatly limits their performance in fine-grained tasks.   

\subsection{Visual Representations for MLLMs}
The pursuit of better visual representations has been a popular research trend in the VL community~\cite{lu2019vilbert,indefense,radford2021learning,ren2024grounded}. Early endeavors mainly explore the object-level features for VL models~\cite{lu2019vilbert,zhang2021vinvl}. Driven by the large-scale image-text pre-training, grid features from CLIP~\cite{radford2021learning} have demonstrated the great efficiency and generalization in MLLMs~\cite{llava,chen2022pali,alayrac2022flamingo}.  Based on grid features, existing researchers mainly improve  visual representations by scaling up the visual encoder. For example, PaLI~\cite{chen2022pali} increases the parameters of visual encoder  to 3 billions and shows the significant performance boost of MLLMs.  In contrast to these works, we improve the visual representations for MLLMs from the perspective of image resolution, and propose a novel and efficient solution, namely mixture-of-resolution adaptation.

 	\begin{figure*}[t]
		\centering
		\includegraphics[width=1.\textwidth]{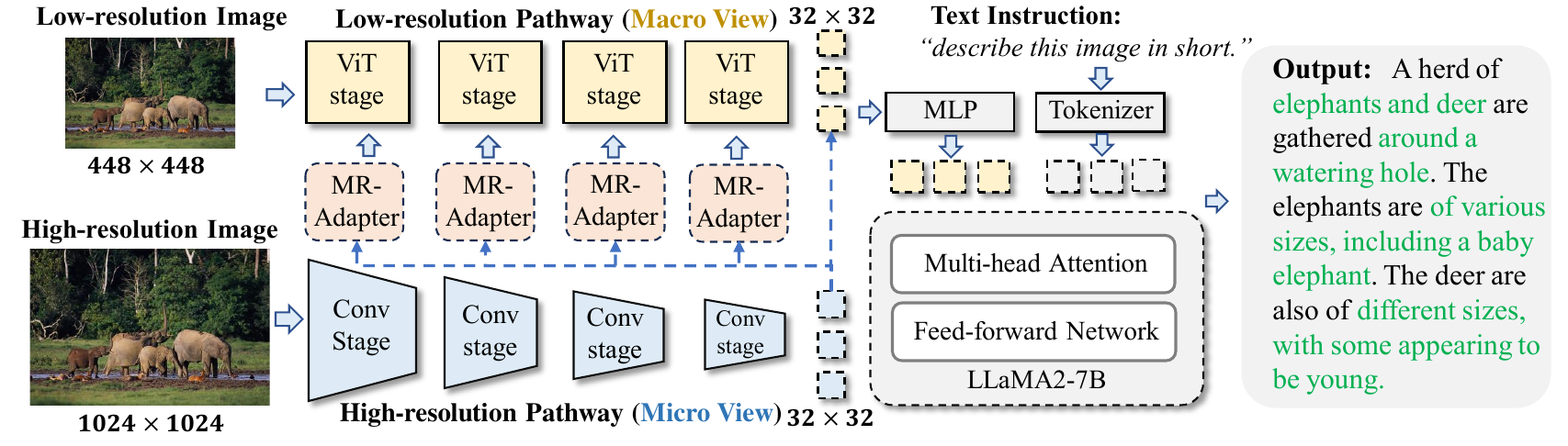} 
  		\vspace{-2em}
		\caption{\textbf{ Illustration of  Mixture-of-Resolution Adaptation (MRA) and its deployment on LLaVA-HR.}   MRA employs dual visual pathways to process high-resolution and low-resolution images, respectively.  High-resolution information is embeded into the fast pathway via a novel mixture-of-resolution adapter (MR-Adapter).  }
		\label{fig2}
		\vspace{-1em}
	\end{figure*}

\section{Preliminary}
\label{sec:limitation}
We first recap the structure of multimodal large language models (MLLMs), which  consists of   an image encoder $\mathcal{F_I(\cdot)}$, an intermediate network $\mathcal{F_P(\cdot)}$ and an LLM $\mathcal{F_{L}(\cdot)}$. 

In particular, given an input image $I \in \mathbb{R}^{H \times W \times 3}$ and a textual instruction $T \in \mathbb{R}^{L}$,  the visual tokens $\textbf{F}_v \in \mathbb{R}^{ (h \times w) \times d}$ are obtained via the image encoder,  and the text tokens $f_t \in \mathbb{R}^{ l \times d}$ are represented by  the  corresponding word embeddings.  Based on the visual and textual  tokens, the LLM will decode the  target  word step by step,   formulated  as
\begin{equation}
\begin{aligned}
p_t=\prod_{s=1}^{S+1}\mathcal{F_{L}}(R_s|\mathcal{F_P}(\textbf{F}_v),f_t,R_{0:s-1}).
\end{aligned}
\label{eq_mllm}
\end{equation}
Here,  $p_t\in \mathbb{R}^{m}$ denotes the  probabilities of the predicted word and $m$ is the size of word vocabulary.

In some MLLMs~\cite{llava,llava1.5}, $\mathcal{F_P}(\cdot)$ is often a stack of simple linear  layers, which are used to directly project the visual tokens onto the semantic space of LLMs. Although simple and effective, this strategy inevitably leads to a longer visual sequence as the resolution increases, \emph{e.g.,} 5,329 tokens for 1,022 $\times$ 1,022 resolution in LLaVA-1.5.   In practice, processing such a large number of tokens is computationally expensive in MLLMs. 
To further  reduce the number of visual tokens, recent advances adopt the sampler-based module for \textbf{ $\mathcal{F_P}(\cdot)$ }, \emph{e.g.,} \textit{QFormer}~\cite{li2023blip}, which  aggregates visual features into several tokens that LLM can directly handle.  Nevertheless, these methods often require large-scale pre-training to achieve VL alignments~\cite{bai2023qwen,li2023blip}. 

Based on the above analyses, we conclude that the main difficulty of high-resolution  image adaptation lies in the rapidly growing visual sequence. This  issue  motivates  us to further explore how to efficiently encode richer visual information with fewer visual tokens.

\section{Mixture-of-Resolution Adaptation}

\subsection{Overview}
To address the above issues, we propose a novel  and efficient method for MLLMs,   termed \textit{mixture-of-resolution adaptation} (MRA), of which structure is depicted in Fig.~\ref{fig2}. The core idea of MRA is to embed high-resolution information into the low-resolution  one via a dual pathway design.  In this case, MRA can keep  a smaller number of visual tokens while encoding richer visual information.  

Particularly, given the input images of two resolutions $I_{l} \in \mathbb{R}^{H_l\times W_l \times 3}$ and  $I_{h} \in \mathbb{R}^{H_h\times W_h \times 3}$, the process of MRA can be formulated as
	\begin{equation} 
	\begin{aligned} 
 &\textbf{F}_v=\mathcal{F}_{\mathcal{I}_l}(I_l,\mathcal{F_{A}}(\textbf{F}_{vh} )) + \textbf{F}_{vh},\\
 &\textbf{F}_{vh}=\mathcal{F}_{\mathcal{I}_h}(I_h).
	\end{aligned}
 \label{eq_framework}
	\end{equation}
Here,  $\textbf{F}_{vh} \in \mathbb{R}^{h_h\times w_h \times d_h}$ and $\textbf{F}_v \in \mathbb{R}^{h\times w \times d}$ denote the high-resolution features and the final visual features, respectively.
And $\mathcal{F}_{\mathcal{I}_l}$ and $\mathcal{F}_{\mathcal{I}_h}$ are the visual encoders for high-resolution and low-resolution images, respectively. $\mathcal{F_{A}}$ denotes the \textit{mixture-of-resolution adapter} (MR-Adapter). In Eq.~\ref{eq_framework}, MRA adopts dual visual pathways to process high- and low- resolution images  simultaneously. Then, a novel MR-Adapter is used to fuse the high-resolution information from the slow pathway  to  the fast one. Finally, the  visual features of two resolutions are  combined  and processed by the LLM based on Eq.~\ref{eq_mllm}.

	\begin{figure}[t]
		\centering
		\includegraphics[width=0.32\textwidth]{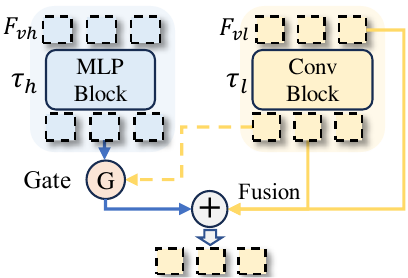} 
  		\vspace{-1em}
		\caption{\textbf{Illustration of the mixture-of-resolution adapter (MR-Adapter).}   MR-Adapter can dynamically  embed the high-resolution features into the low-resolution pathway.  }
		\label{fig3}
		\vspace{-1em}
	\end{figure}
\subsection{Dual Visual Pathways}

As shown in Fig.~\ref{fig2}, dual visual pathways are the key design of MRA, and their benefits are maximized   from two aspects.

\textbf{Visual functionality.} Firstly, the dual visual pathways  process images from macro- and micro-views, which is inspired by the visual system of human being~\cite{merigan1993parallel,robertson1991neuropsychological}.
Particularly, \citet{robertson1991neuropsychological} find that the visual system processes local and global semantics  via different pathways.   Based on this finding, we adopt a similar mechanism to our  MRA. Specifically, one  visual  pathway aims to capture fine-grained semantics from high-resolution images  \emph{i.e.}, processing  images  from  local view.  In contrast,  the other   pathway is designed to encode global information from low-resolution images, achieving  a larger receptive field.

\textbf{Visual alignment.} Due to different resolutions, these two pathways often  produce  visual features of different shapes, impeding their quick  alignments~\cite{yu2019multimodal}. To overcome  this limitation,  we adopt  different  downsampling rates for  the low- and high-resolution pathways, respectively.  Thus, their output features can  keep  the same spatial shape. 

Based on the above observations,  we design the dual visual pathways   with  a convolutional network (CNN)~\cite{convnext}  and a vision transformer (ViT)~\cite{dosovitskiy2020image}.  Specifically, CNN is equipped with a downsampling stride of 32 to  process high-resolution images.  ViT encodes low-resolution images with a downsampling stride of 14.  Notably, such designs also ensure the efficiency of MLLMs, where the high-resolution images are processed by the efficient CNN, and the number of visual tokens is also kept small via the large downsampling stride.

\begin{table*}[t]
\caption{\textbf{Performance and efficiency comparisons of LLaVA-HR and LLaVA-1.5~\cite{llava1.5} at different resolutions.}  Except   resolution,  the other  configurations of LLaVA-HR and LLaVA-1.5 remain the same.  The training and inference costs are measured on  NVIDIA A800s.  ``\textit{N/A}'' denotes that  GPU memory overflows\footnotemark[1].   ``\textit{tokens/s}'' denotes the number of generated tokens per second. }
\vspace{.5em}
\centering
		\renewcommand\arraystretch{1.2}
		\setlength\tabcolsep{5pt}
\begin{tabular}{lc|cccc|ccc}
\toprule
\multicolumn{1}{c}{\multirow{2}{*}{Models}}&\multicolumn{1}{c|}{\multirow{2}{*}{Resolution}} & \multicolumn{4}{c|}{Vision-Language Tasks} & \multirow{2}{*}{\begin{tabular}[c]{@{}c@{}}Training \\ Time $\downarrow$\end{tabular}} & \multirow{2}{*}{\begin{tabular}[c]{@{}c@{}}GPU\\ Memory $\downarrow$\end{tabular}} & \multirow{2}{*}{\begin{tabular}[c]{@{}c@{}}Inference\\ Speed $\uparrow$\end{tabular}} \\
          &   \multicolumn{1}{c|}{}          & VQAv2 $\uparrow$   & TextVQA  $\uparrow$  & MME   $\uparrow$    & POPE  $\uparrow$ &                                                                           &                                                                       &                                                                            \\ \hline
LLaVA-1.5~& 336 pix                           & 80.44    & 59.41      & 1461.17   & 86.2   & 15.6h                                                                     & 28G                                                                   & 23.8 tokens/s                                                                   \\
\rowcolor{lightgray} LLaVA-HR (ours) & 384 pix                     & 80.47    & 59.63      & 1522.28   & 86.3   & 17.6h                                                                     & 34G                                                                   & 23.8 tokens/s                                                                   \\ \hline
LLaVA-1.5& 448 pix                           & 81.17    & 62.17      & 1493.12   & 87.2   & 19.4h                                                                     & 49G                                                                   & 19.9 tokens/s                                                                   \\
\rowcolor{lightgray} LLaVA-HR (ours) & 768 pix                   & 81.80    & 64.36      & 1524.75   & 88.0   & 18.2h                                                                     & 38G                                                                   & 23.5 tokens/s                                                                   \\ \hline
LLaVA-1.5& 672 pix                           & 81.54    & 64.23      & 1498.71   & 87.9   & 31.8h                                                                     & 79G                                                                   & 12.7 tokens/s                                                                   \\
\rowcolor{lightgray} LLaVA-HR (ours)& 1024 pix                    & 81.90    & 67.11      & 1554.90   & 87.6   & 20.7h                                                                     & 40G                                                                   & 19.7 tokens/s                                                                   \\ \hline
LLaVA-1.5& 1022 pix                          &      74.20    &     37.80       &     1266.90      &    84.4    & 69.4h                                                                     & N/A\footnotemark[1]                                                                   & 5.6 tokens/s                                                                    \\
\rowcolor{lightgray} LLaVA-HR (ours)& 1536 pix                      & 81.82    & 67.96      & 1480.62   & 87.7   & 29.8h                                                                     & 52G                                                                   & 12.6 tokens/s                                                                   \\ \bottomrule
\end{tabular}
\label{tab1}
\vspace{-.5em}
\end{table*}

\subsection{Mixture-of-Resolution Adapter}
To better  collaborate the  feature learning  of two pathways, we propose a \textit{mixture-of-resolution adapter} (MR-Adapter)  for the fusion of visual information from different resolution images. In particular, given the  visual features  $\textbf{F}_{vh} \in  \mathbb{R}^{h\times w \times d_h} $ extracted from a high-resolution image, we  embed  them  into the low-resolution visual pathway   by
	\begin{equation} 
	\begin{aligned} 
 \textbf{F}_{vl}'= \textbf{F}_{vl} + f_l(\textbf{F}_{vl} )+ g \cdot f_h(\textbf{F}_{vh} ).
	\end{aligned} 
 \label{adapter}
	\end{equation}
Here, $\textbf{F}_{vl} \in  \mathbb{R}^{h\times w \times d_l}$  are  the features from the low-resolution pathway.  $f_l(\cdot)$ and  $f_h(\cdot)$ denote two mapping modules, which are designed as   a convolutional block and  an MLP layer, respectively. $g$ is a dynamic score to control the weights of high-resolution information, defined by
	\begin{equation} 
	\begin{aligned} 
  g &=\delta(W_2\sigma(W_1f_v)),\\
 f_v &=  \frac{1}{h\times w}\sum_i^{h}\sum_j^{w} [f_l(\textbf{F}_{vl} )^{i,j}, f_h(\textbf{F}_{vh} )^{i,j}].
	\end{aligned} 
	\end{equation} 
Here, $[\cdot] $ denotes the concatenation operation, and $W_1\in \mathbb{R}^{2d\times\frac{d}{2}}$ and $W_2\in \mathbb{R}^{\frac{d}{2}\times d}$ are two projection  matrices. $f_v \in \mathbb{R}^{d}$ is the  pooled visual features.  $\sigma$ and $\delta$ denote the activation function of \textit{GELU} and \textit{Tanh}, respectively.

As shown in Fig.~\ref{fig2},   high-resolution information can be fused  with the features in each block of  ViT.    In this case, the low-resolution features  of ViT also contain rich   semantics,  improving the visual  descriptive power  of MLLMs. 


\subsection{The Deployment on MLLM}
We apply MRA to a popular MLLM  called LLaVA-1.5~\cite{llava1.5}, and construct a new model, namely LLaVA-HR.   Its training  consists of two stages, \emph{i.e.},  low-resolution pre-training and high-resolution instruction tuning.

\textbf{Stage 1: Low-Resolution Pre-training.}  Similar to LLaVA~\cite{llava} and LLaVA-1.5~\cite{llava1.5}, this stage aims to optimize the projector to align the visual features with the word embeddings of   LLM. Therefore, the image encoder and the LLM are   frozen during  pre-training. Besides, we adopt  low resolutions for two pathways. In this stage, the MR-Adapter is not inserted, and  output features of dual pathways are  directly combined.

\textbf{Stage 2: High-Resolution Instruction Tuning.}
During instruction tuning, we greatly increase the resolution of the high-resolution pathway, \emph{e.g.,} from 384$\times$ 384 to 1,024$\times$ 1,024. And  the low-resolution one is also accordingly adjusted to ensure the visual alignment of two pathways, \emph{e.g.,} from 336$\times$ 336 to 448$\times$ 448. Meanwhile, the MR-Adapter is  then applied to  connect two visual pathways.  Different from the first  training  stage, the entire MLLM will be  fully optimized   to better accommodate high-resolution images.

\footnotetext[1]{ When memory overflows, we reduce the batch size and increase the gradient accumulation  steps to train LLaVA-1.5.}

\section{Experiments}
\subsection{Evaluations and Metrics}
\textbf{Multimodal benchmarks for MLLM.} We evaluate LLaVA-HR on four  emerging multimodal benchmarks for MLLMs, including MME~\cite{fu2023mme}, POPE~\cite{li2023pope}, SEED~\cite{li2023seed} and MM-VET~\cite{yu2023mmvet}.  In particular, MME and MM-VET evaluate the multimodal perception and cognition abilities of MLLMs. SEED extends the modalities of evaluation  to  images and videos.  POPE aims to evaluate the visual hallucinations of MLLMs. The metrics used in our paper follow  their default settings. For MME, we follow LLaVA-1.5 to report the  perception score.

\textbf{Common vision-language benchmarks.}  We  also evaluate LLaVA-HR on seven VL datasets, including VQAv2~\cite{goyal2017vqav2}, GQA~\cite{hudson2019gqa}, OKVQA~\cite{okvqa}, OCRVQA~\cite{mishra2019ocrvqa}, ScienceQA~\cite{lu2022learn}, VizWiz~\cite{gurari2018vizwiz} and TextVQA. In particular,   ScienceQA~\cite{lu2022learn}, VizWiz~\cite{gurari2018vizwiz} and TextVQA are three \textbf{zero-shot tasks}, and their samples are not appeared in our training data. We report the accuracy on the \textit{test} set of OCRVQA, the \textit{test} set of VizWiz, and the \textit{val} set of OKVQA.  We organize samples of these tasks in instruction formats of LLaVA-1.5~\cite{llava1.5}.

\subsection{Implementation Details} 
In LLaVA-HR, we use CLIP-ViT-L~\cite{radford2021learning,openclip} and CLIP-ConvNeXt-L~\cite{convnext} as the dual visual paths to encode low- and high-resolution images, respectively.  In {LLaVA-HR-X}, the CLIP-ConvNeXt-L is replaced with the stronger CLIP-ConvNeXt-XXL. The MR-Adapter is applied into the last three stages of ViT. 
Following LLaVA-1.5, we first pre-train LLaVA-HR on LCS-558K~\cite{llava}, which contains 558\textit{k} image-text pairs. During the  pre-training stage, both the visual encoder and the LLM are frozen, and only the MLP projector is fine-tuned.   AdamW~\cite{kingma2014adam} is used as the optimizer, and the learning rate and batch size are set to 1e-3 and 256, respectively.   Visual resolutions are set to 336$\times$336 and 384$\times$384 for the ViT and the CNN, respectively.  During  instruction tuning,  we  follow LLaVA-1.5 to use 665\textit{k} VL instruction data. At this stage, the entire model is updated with a learning rate of 2e-5. Besides, we increase the resolution of   ViT and   CNN to 448$\times$448 and 1,024$\times$1,024, respectively.  The training epoch is set to 1 for  pre-training and instruction tuning.

\begin{table}[t]
		\renewcommand\arraystretch{1.2}
		\setlength\tabcolsep{1 pt}
  \caption{\textbf{Comparison of MRA and four baselines on LLaVA-HR.} The visual resolution is set to about $\sim$760$\times$ 760.  }
\begin{tabular}{l|cccc|c}
\toprule
Settings             & VQAv2 & TextVQA & MME    & POPE &Speed \\ \midrule
ViT+ MLP         & 81.0   &  63.2   &  1436.1 &   87.6   &10.7  t/s       \\
Conv+MLP         & 80.3  & 64.6    & 1415.9 & 86.6     &23.7 t/s            \\
ViT+Resampler   &  79.8     &     58.9    &  1403.8      &  85.8&      27.6 t/s        \\
ViT+Pooling+MLP &   80.6    &   59.6      &   1480.6     &   86.5 &     23.9 t/s                 \\
\rowcolor{lightgray}MRA (ours) &    81.8  & 64.4    & 1524.8 & 88.0  &23.5 t/s           \\ \bottomrule
\end{tabular}
\label{tab2}
\vspace{-1.5em}
\end{table}

\begin{table}[t]
\caption{\textbf{Ablation study of mixture-of-resolution adaptation on LLaVA-HR.}  The  resolution is 768 $\times$ 768.  Our final setting  is colored in gray. ``L-Res Path.'', ``H-Res Path.'', ``Fusion Direct.'', ``Struct.'' and ``Gate Fuct.'' denote the low-resolution pathway, the high-resolution pathway, the fusion direction, the structure and the gate function, respectively.  }
\vspace{.5em}
\centering
		\renewcommand\arraystretch{1.2}
		\setlength\tabcolsep{1.4pt}
\begin{tabular}{l|c|cccc}
\toprule
\multicolumn{1}{c|}{Settings}                                             & Choices      & VQAv2 & TextVQA & MME    & POPE \\ \hline
\multirow{3}{*}{\begin{tabular}[c]{@{}l@{}}L-Res\\ Path.\end{tabular}}      &  
   \cellcolor{lightgray} ViT-L        & \cellcolor{lightgray}81.8  &\cellcolor{lightgray}64.4    & \cellcolor{lightgray}1524.8 &\cellcolor{lightgray} 88.0 \\
                                                                          & None         & 80.3  & 64.6    & 1415.9 & 86.6 \\
                                                                          & ViT-G        & 81.7  & 65.3    & 1469.7 & 87.9 \\ \midrule
\multirow{3}{*}{\begin{tabular}[c]{@{}l@{}}H-Res\\ Path.\end{tabular}}      & \cellcolor{lightgray} ConvNeXt-L   &\cellcolor{lightgray} 81.8  & \cellcolor{lightgray}64.4    & \cellcolor{lightgray}1524.8 & \cellcolor{lightgray}88.0 \\
                                                                          & None         & 80.4  & 59.4    & 1461.2 & 86.2 \\
                                                                          & ConvNeXt-XXL & 82.3  & 66.5    & 1479.2 & 87.9 \\ \midrule
\multirow{2}{*}{\begin{tabular}[c]{@{}l@{}}Fusion \\ Direct.\end{tabular}} & \cellcolor{lightgray}High to Low & \cellcolor{lightgray}81.8  & \cellcolor{lightgray}64.4    & \cellcolor{lightgray}1524.8 & \cellcolor{lightgray}88.0 \\
                                                                          & Low  to High &      81.0 &    62.8     &      1463.5  &  87.3    \\ \midrule
\multirow{2}{*}{\begin{tabular}[c]{@{}l@{}}Fusion \\ Type\end{tabular}}   & \cellcolor{lightgray}Sum          & \cellcolor{lightgray}81.8  & \cellcolor{lightgray}64.4    & \cellcolor{lightgray}1524.8 & \cellcolor{lightgray}88.0 \\
                                                                          & Concat       & 81.7  & 64.7    & 1508.8 & 87.3 \\ \midrule
\multirow{3}{*}{Struct.}                                                    & \cellcolor{lightgray}mlp-conv     & \cellcolor{lightgray}81.8  & \cellcolor{lightgray}64.4    & \cellcolor{lightgray}1524.8 &\cellcolor{lightgray} 88.0 \\
                                                                          & conv-conv    & 81.6  & 64.6    & 1499.0 & 87.7 \\
                                                                          & conv-mlp     & 81.5  & 64.2    & 1517.9 & 87.6 \\ \midrule
\multirow{3}{*}{\begin{tabular}[c]{@{}l@{}}Gate \\ Funct.\end{tabular}}    & \cellcolor{lightgray}Tanh         & \cellcolor{lightgray}81.8  & \cellcolor{lightgray}64.4    & \cellcolor{lightgray}1524.8 & \cellcolor{lightgray}88.0 \\
                                                                          & Sigmoid      & 81.7  & 64.3    & 1567.9 & 86.9 \\
                                                                          & H-sigmoid    & 81.6  & 64.4    & 1525.9 & 87.8 \\ \bottomrule
 
\end{tabular}
\label{tab3}
\vspace{-1em}
\end{table}


\begin{table*}[t]
\caption{\textbf{Comparison with existing methods on four MLLM benchmarks.}  ``Param.'', ``Res.'' and ``Data'' refer to the total parameters, the visual resolution and the number of training data, respectively. ``t/s'' refers to tokens per second. }
\vspace{.5em}
\centering
		\setlength\tabcolsep{8pt}
\begin{tabular}{llcl|cccc|c}
\toprule
\multirow{2}{*}{Method} & \multicolumn{3}{c|}{Settings}                                                     & \multicolumn{4}{c}{Multimodal Benchmarks}  & Inference                     \\
                        & \multicolumn{1}{c}{Param.} & \multicolumn{1}{c}{Res.} & \multicolumn{1}{c|}{Data} & MME             & POPE          & SEED          & MM-Vet  &Speed      \\ \midrule
BLIP-2                  & 14.2B                      & \multicolumn{1}{c}{224}  & \multicolumn{1}{c|}{129M} & 1293.8          & 85.3          & 46.4          & 22.4   &-       \\
InstructBLIP            & 8.2B                    & \multicolumn{1}{c}{224}  & \multicolumn{1}{c|}{130M} & -               & -             & 53.4          & 26.2    & 22.6 t/s         \\
InstructBLIP            & 14.2B                      & \multicolumn{1}{c}{224}  & \multicolumn{1}{c|}{130M} & 1212.8          & 78.9          & -             & 25.6    &-      \\
QWen-VL-Chat            & 9.6B                       & \multicolumn{1}{c}{448}  & \multicolumn{1}{c|}{1.4B} & 1487.5          & -             & 58.2          & -     &17.0 t/s        \\
Fuyu-8B & 8B &$\sim$600 & - & 728.6&74.1&-&21.4 &15.6 t/s\\ 
mPLUG-Owl2&8.2B&448&400M&1450.2&-&57.8&\textbf{36.2}& 19.6 t/s\\
LLaVA-1.5               & 7.2B                       & 336                      & 1.2M                      & 1510.7          & 85.9          & 58.6          & 30.5   &23.8 t/s       \\
LLaVA-1.5               & 13.2B                      & 336                      & 1.2M                      & 1531.3          & 85.9          & 61.6          & 35.4   &-       \\ \midrule
\rowcolor{lightgray}LLaVA-HR                & 7.4B                       & 1024                     & 1.2M                      & \textbf{1554.9} & 87.6          & 64.2          & 31.2    &19.7 t/s      \\
\rowcolor{lightgray}LLaVA-HR                & 13.4B                      & 1024                     & 1.2M                      & 1540.9          & 87.8          & 64.5          & 34.8    & 15.0 t/s      \\
\rowcolor{lightgray}LLaVA-HR-X              & 14B                        & 1024                     & 1.2M                      & 1487.3          & \textbf{88.0} & \textbf{65.3} & {35.5} &12.9 t/s\\ \bottomrule
\end{tabular}
\vspace{-1 em}
\label{tab4}
\end{table*}

\begin{table*}[t]
\caption{\textbf{Comparison with existing methods on seven vision-language tasks.} SQA$^I$  refers to the \textit{IMG} subset of ScienceQA.    } 
\vspace{.5em}
\centering
		\setlength\tabcolsep{4pt}
\begin{tabular}{llll|cccc|ccc|c}
\toprule
\multirow{2}{*}{Method} & \multicolumn{3}{c|}{Settings}                                                  & \multicolumn{4}{c|}{In-domain Tasks}                            & \multicolumn{3}{c}{Zero-shot Tasks}    & Infer.       \\
                        & \multicolumn{1}{c}{Param.} & \multicolumn{1}{c}{Res.} & \multicolumn{1}{c|}{Data} & VQAv2 & GQA           & OKVQA         & OCRVQA        & SQA$^I$       & VizWiz        & TextVQA  &Speed      \\ \midrule
BLIP-2 & 14.2B & 224 & 129M  & 41.0            & 41.0          & 45.9          & 40.6          & 61.0          & 19.6          & 42.5   &-       \\
InstructBLIP & 8.2B & 224 & 130M  & -               & 49.2          & -             & -             & 60.5          & 34.5          & 50.1   & 22.6  t/s       \\
InstructBLIP & 14.2B & 224 & 130M    & -               & 49.5          & -             & 44.8          & 63.1          & 33.4          & 50.7  &-        \\
Shikra & 13.2B & 224 & 6.1M    & 77.4            & -             & -             & -             & -             & -             & -   &-          \\
IDEFICS-9B & 9B & 224 & 354M                           & 50.9            & -             & 38.4          & -             & -             & 35.5          & 25.9    &30.5  t/s           \\
IDEFICS-80B & 80B & 224 & 354M                           & 60.0            & -             & 45.2          & -             & -             & 36.0          & 30.9   &-  \\
Qwen-VL-Chat & 9.6B & 448 & 1.4B                          & 78.2            & 57.5          & 56.6          & \textbf{70.5} & 68.2          & 38.9          & 61.5    &17.0  t/s      \\ 
Fuyu-8B & 8B &$\sim$600& -& 74.2&-&60.6&-&-&-&-&15.6  t/s\\
mPLUG-Owl2&8.2B&448&400M&79.4& 56.1& 57.7&-&68.7&54.5&58.2&19.6 t/s \\
LLaVA-1.5 & 7.2B & 336 & 1.2M                           & 78.5            & 62.0          & -             & -             & 66.8          & 50.0          & 58.2   &23.8  t/s       \\
LLaVA-1.5 & 13.2B & 336 & 1.2M                             & 80.0            & 63.3          & -             & -             & \textbf{71.6} & 53.6          & 61.3        &-  \\  \midrule 
\rowcolor{lightgray}LLaVA-HR                & 7.4B               & 1024                     & 1.2M                      & 81.9            & 64.2          & 58.9          & 68.4          & 65.1          & 48.7          & 67.1   &19.7  t/s       \\
\rowcolor{lightgray}LLaVA-HR                & 13.4B              & 1024                     & 1.2M                      & 82.3              & 64.8          & 60.7          & 67.7          & 68.1          & \textbf{57.9 }         & 68.1   &15.0 t/s       \\
\rowcolor{lightgray}LLaVA-HR-X              & 14B              & 1024                     & 1.2M                      & \textbf{82.6}   & \textbf{65.2} & \textbf{61.5} & 69.0          & 68.0          & {56.6} & \textbf{70.9} &12.9  t/s \\ \bottomrule
\end{tabular}
\label{tab5}
\vspace{-1.em}
\end{table*}
\subsection{Experimental Results}
\subsubsection{Quantitative  Analysis}
\textbf{Comparison with baselines.}
In Tab.~\ref{tab1}, we compare the performance and efficiency of  LLaVA-HR with   LLaVA-1.5~\cite{llava1.5}   with different  image resolutions. From this table, we observe that increasing  image  resolution obviously improves the performance of two models on four tasks, \emph{e.g.,} +4.8\% of LLaVA-1.5 on TextVQA.  However,  the performance of  LLaVA-1.5 drops significantly at the resolution of 1,024$\times$1,024. To explain, the number of visual tokens greatly exceeds the pre-trained context length of the LLM, which easily causes the instability during training. In contrast, the performance of LLaVA-HR is consistently improved from 384 $\times$ 384 resolution to 1,024 $\times$ 1,024 resolution. Besides, the total gain of LLaVA-HR is more obvious than that of LLaVA-1.5~\cite{llava1.5}, \emph{e.g.,} +8.33\% of LLaVA-HR \textit{vs.} +4.82\% of LLaVA-1.5,  greatly confirming the effectiveness of MRA.

In Tab.~\ref{tab2}, we further compare four common baselines with the similar resolution, \emph{i.e.,} $\sim$760$\times$760. ``ViT+MLP'' is the default  setting of  LLaVA-1.5 as the reference. ``Conv+MLP'' replaces the visual backbone with  ConvNeXt~\cite{convnext}, which uses a larger downsampling rate to reduce the number of visual tokens.  ``ViT+Resampler''  and ``ViT+Pooling+MLP'' refer to  the two pooling strategies  for reducing the number of visual tokens.   As can be seen,  all compared methods are inferior to LLaVA-HR. In particular, using a convolutional network as the visual backbone greatly improves efficiency, but its performance still lags behind LLaVA-HR by a large margin, \emph{e.g.,} -108.9 on MME~\cite{fu2023mme}. Similarly, ``ViT+Resampler''  and ``ViT+Pooling+MLP'' also sacrifice performance for efficiency.   Overall, these comparisons further confirm the designs of MRA.

Despite  effectiveness, the expenditure of LLaVA-HR is  also cost-effective. In particular, increasing resolution from 384 $\times$ 384 to 1,024 $\times$ 1,024  slows down the training and inference of  LLaVA-1.5 by 344.8\% and 325\%, respectively. However, these   costs are  reduced to only  17.6\% and 20.8\%  in  LLaVA-HR.   Despite   better performance, the training   and inference speeds of LLaVA-HR are three times faster than LLaVA-1.5.    Besides, the costs of GPU memory also remain cheap for LLaVA-HR. For example, adapting the resolution of 1,536 $\times$ 1,536 for LLaVA-HR only consumes 52G GPU memory, but the same settings for LLaVA-1.5 will cause  GPU memory overflow.  These results greatly confirm the  efficiency of our MRA and LLaVA-HR.

	\begin{figure*}[t]
		\centering
		\includegraphics[width=.97\textwidth]{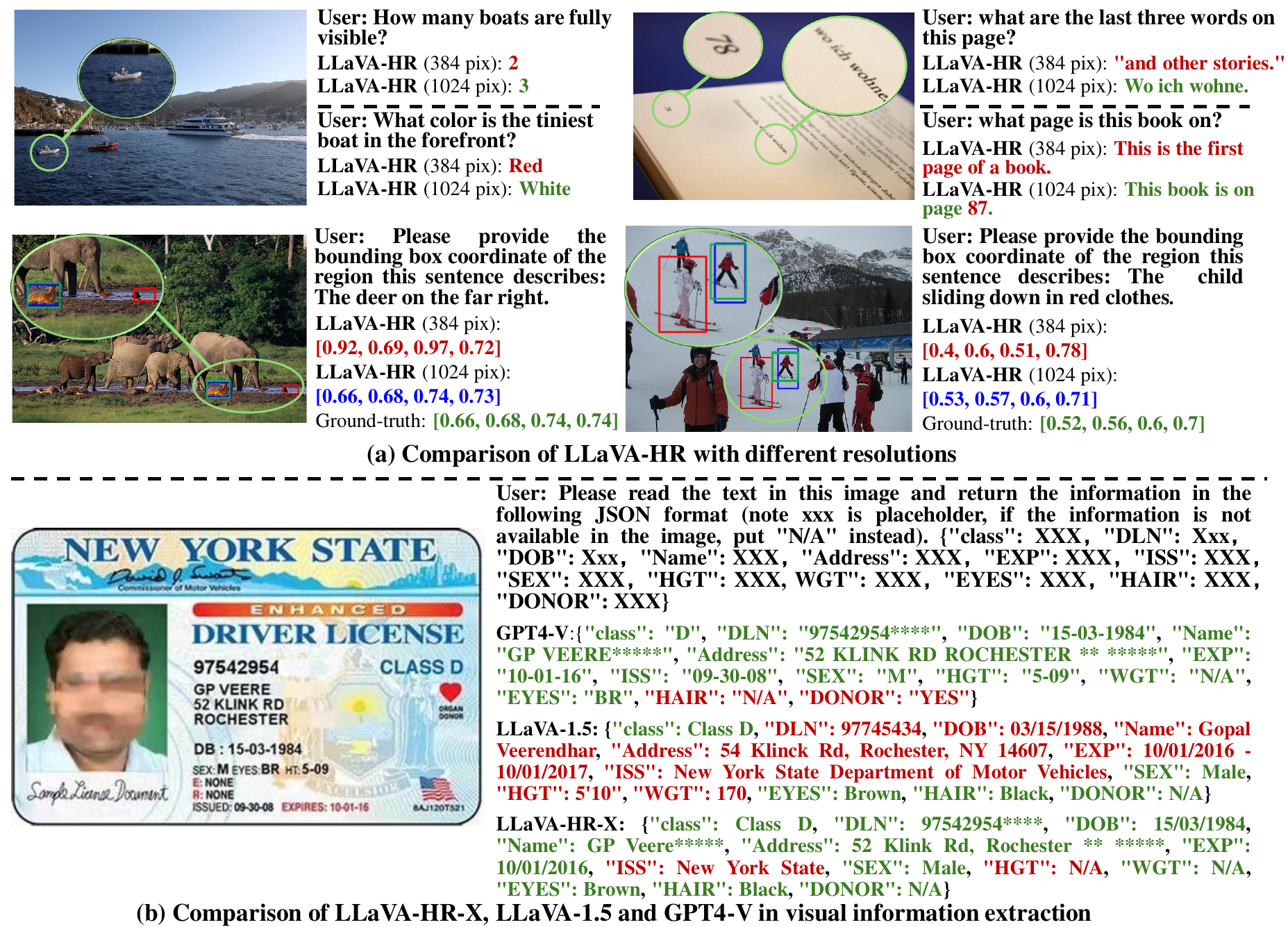} 
  		\vspace{-1em}
		\caption{\textbf{Visualizations of LLaVA-HR and existing MLLMs.}   Subfig-(a) shows  that  high image resolution greatly improves the capability of MLLMs on fine-grained  VL  tasks.  In  Subfig-(b), LLaVA-HR-X demonstrates the comparable  ability  with GPT4-V in visual information extraction\footnotemark[2].    Correct and  incorrect answers are colored in green and red, respectively.}
		\label{fig6} 
  \vspace{-1em}
	\end{figure*}

\textbf{Ablation studies.}
In Tab.~\ref{tab3}, we conduct comprehensive ablation studies for MRA on four VL benchmarks. Firstly, we validate the different designs of the dual visual pathways. From  these  results, we find that removing one  pathway will lead to significant performance drops, \emph{e.g.,} -1.5\% on VQAv2.   Besides,  scaling up  the  high-resolution  encoder brings more gains than that of the low-resolution one, \emph{e.g.,} +2.1\% \textit{vs.} +0.9\% on TextVQA. We assume that the stronger high-resolution image encoder can better capture the fine-grained visual information. Then, we  ablate different fusion directions and strategies in MRA. Specifically,   changing  the fusion direction  obviously degenerates the performance, \emph{e.g.,} -61.3  on MME.      Finally, we ablate the designs of the mixture-of-resolution adapter.  Specifically, the best choices of mapping modules for the low- and high-resolution pathways are convolution blocks  and MLP blocks, respectively.  Besides, the choices of gating function also affect performance and the \textit{tanh} function perform the best. These ablations further confirm  the designs of MR-Adapter.

\textbf{Comparison with existing MLLMs.}
In Tab.~\ref{tab4} - \ref{tab5}, we compare LLaVA-HR with existing MLLMs on 11 VL tasks.   On  the  four MLLM benchmarks, we observe  comprehensive advantages of LLaVA-HR against existing MLLMs. In particular, LLaVA-HR achieves 1554.9 scores in MME benchmark, outperforming LLaVA-1.5  by +23.6. On POPE, a benchmark including  video evaluations,  LLaVA-HR-X still outperforms existing MLLMs by a large margin, \emph{i.e.,} +3.7\% gains. Besides, LLaVA-HR achieves the best performance on the benchmark  for   visual hallucinations, \emph{i.e.,} POPE, suggesting  that its  visual hallucinations are greatly alleviated. Notably, Fuyu-8b~\cite{fuyu} is  capable  of high-resolution images, but its performance is  much inferior to LLaVA-HR, \emph{e.g.,} 728.6 \textit{vs.} 1554.9 on MME.

Tab.~\ref{tab5}  gives the   performance comparison  on  common VL tasks.  On  in-domain tasks, LLaVA-HR achieves the best results on three tasks, \emph{e.g.,} 82.6 on VQAv2 and 61.5 on OKVQA. On OCRVQA, Qwen-VL-Chat  collects more in-domain data for training, so it performs  better than LLaVA-HR.   Under the  zero-shot  setting, we can observe  more significant advantages of LLaVA-HR  on the fine-grained tasks, \emph{e.g.,} VizWiz and TextVQA.  Most notably, even Qwen-VL-Chat is pre-trained with 24.8M OCR  samples, it still performs worse than LLaVA-HR-X on TextVQA. These results suggest the significance of high resolution for these tasks. In contrast, most images of ScienceQA are synthetic and of low resolution, so the advantages of LLaVA-HR are not obvious. Overall, these results greatly confirm the effectiveness and generalization of LLaVA-HR and our MRA.

\subsubsection{Qualitative Experiments}
  \footnotetext[2]{For privacy reasons, we blur some key personal information.} 
 In   Fig~\ref{fig6} (a), we compare the predictions of LLaVA-HR with different resolutions.   The visualizations show that higher image resolution  obviously improves the capability of MLLMs  on  fine-grained tasks. For example, LLaVA-HR with a resolution of 1,024 $\times$ 1,024 can well capture  granular visual content, \emph{e.g.,} the tiny boat in the first example. Besides,  high image resolution also enables LLaVA-HR a 
 stronger ability of text recognition. For instance, the small and  blurred phrase  of ``\textit{wo ich wohne}'' in the second example are correctly identified by the high-resolution LLaVA-HR.  These results greatly confirm the significance of high image resolution in addressing visual shortcoming.
  In Fig~\ref{fig6}  (b), we further compare the predictions of LLaVA-HR-X, LLaVA-1.5~\cite{llava1.5} and GPT4-V~\cite{gpt4v} in visual information extraction.  Notably, LLaVA-HR-X shows a comparable  ability with GPT4-V on this challenging task. As shown in  Fig~\ref{fig6} (b), LLaVA-HR-X and GPT4-V can correctly extract almost all visual content of the driver  license and organize it in JSON format. Compared to GPT4-V, LLaVA-HR-X  also correctly identifies the hair color of the person, which requires fine-grained visual reasoning. In contrast,  LLaVA-1.5 can only recognize simple visual content like ``\textit{class}'' and ``\textit{SEX}'', and fail to extract most visual information. These results  further   validate the effectiveness of MRA in addressing visual shortcoming of MLLMs.

\section{Conclusion}
In this paper, we  study the visual shortcoming of MLLMs from the perspective of image resolution, and  propose a novel and efficient  method  for high-resolution adaptations of MLLMs, namely \textit{mixture-of-resolution adaptation} (MRA). MRA adopts dual visual pathways to process  images of both  high and low resolutions, where  high-resolution information is embeded into the low-resolution modeling via the novel  \textit{mixture-of-resolution adapters} (MR-Adapters).  We apply MRA to a popular MLLM  called  LLaVA-1.5, and construct a new high-resolution MLLM, termed  LLaVA-HR. Experimental results not only validate the effectiveness  of LLaVA-HR in addressing visual shortcoming, but also confirm its remarkable efficiency against existing MLLMs.

\paragraph{Acknowledgements.}
This work was supported by National Key R\&D Program of China (No.2022ZD0118201) , the National Science Fund for Distinguished Young Scholars (No.62025603), the National Natural Science Foundation of China (No. U21B2037, No. U22B2051, No. 62176222, No. 62176223, No. 62176226, No. 62072386, No. 62072387, No. 62072389, No. 62002305 and No. 62272401),  the Natural Science Foundation of Fujian Province of China (No.2021J01002,  No.2022J06001), and the China Fundamental Research Funds for the Central Universities (Grant No. 20720220068).


\bibliography{example_paper}
\bibliographystyle{icml2024}

\end{document}